\pdfoutput=1
\documentclass[11pt]{article}

\usepackage{acl}

\usepackage{times}
\usepackage{latexsym}

\usepackage[T1]{fontenc}
\usepackage[utf8]{inputenc}

\usepackage{microtype}

\usepackage{amsmath, amsfonts, mathtools, amssymb}
\usepackage{url}
\usepackage{graphicx}
\usepackage{framed}
\usepackage{array}
\usepackage{tabulary}
\usepackage{tabularx}
\usepackage{multirow}
\usepackage{xcolor}
\usepackage{verbatim}  % code text
\usepackage{booktabs}  % formatting table
\usepackage{hyperref}
\usepackage{tablefootnote}
\usepackage{footnote}
\usepackage{todonotes}
\usepackage{lipsum}
\newcommand\blfootnote[1]{%
	\begingroup
	\renewcommand\thefootnote{}\footnote{#1}%
	\addtocounter{footnote}{-1}%
	\endgroup
}

\title{Style Control for Schema-Guided Natural Language Generation}

\author{
{\bf Alicia Y. Tsai}$^{1*}$,
{\bf Shereen Oraby}$^2$,
{\bf Vittorio Perera}$^2$,
{\bf Jiun-Yu Kao}$^2$, \\
{\bf Yuheng Du}$^2$,
{\bf Anjali Narayan-Chen}$^2$,
{\bf Tagyoung Chung}$^2$,
{\bf Dilek Hakkani-Tur}$^2$ \\
$^1$ University of California, Berkeley \\
$^2$ Amazon Alexa AI\\
{\tt aliciatsai@berkeley.edu} \\
{\tt \{orabys,pererv,jiunyk,yuhendu,}\\
{\tt naraanja,tagyoung,hakkanit\}@amazon.com}
}

\begin{document}
	\maketitle
	\begin{abstract}
		Natural Language Generation (NLG) for task-oriented dialogue systems focuses on communicating specific content accurately, fluently, and coherently. While these attributes are crucial for a successful dialogue, it is also desirable to simultaneously accomplish specific stylistic goals, such as response length, point-of-view, descriptiveness, sentiment, formality, and empathy. In this work, we focus on stylistic control and evaluation for schema-guided NLG, with joint goals of achieving both semantic and stylistic control. We experiment in detail with various controlled generation methods for large pretrained language models: specifically, conditional training, guided fine-tuning, and guided decoding. We discuss their advantages and limitations, and evaluate them with a broad range of automatic and human evaluation metrics. Our results show that while high style accuracy and semantic correctness are easier to achieve for more lexically-defined styles with conditional training, stylistic control is also achievable for more semantically complex styles using discriminator-based guided decoding methods. The results also suggest that methods that are more scalable (with less hyper-parameters tuning) and that disentangle content generation and stylistic variations are more effective at achieving semantic correctness and style accuracy.
		\blfootnote{$^{*}$Work done as an intern at Amazon Alexa AI.}
	\end{abstract}
	
	\section{Introduction}
Natural Language Generation (NLG) for task-oriented dialogue focuses on effectively generating responses based on inputs that are frequently in the form of a structured meaning representation (MR) \citep{moryossef-etal-2019-step, dusek-etal-2018-findings, colin-etal-2016-webnlg, wen-etal-2015-semantically}. Recent work has suggested a schema-guided paradigm for task-oriented dialogue by adding descriptions in natural language form  \citep{lin-etal-2021-leveraging, du-etal-2020-schema, rastogi2019towards, 46223}. Compared to structured MRs, dialogue schemata contain much richer contextual information, leading to better generated outputs. 

Although the primary aim of task-oriented NLG is to effectively generate outputs that realize system dialogue actions and communicate their associated contents correctly, it is often desirable to control the stylistic variations of an output. For example, recognizing and reacting to emotions has been shown to enhance task outcomes and user engagement in task-oriented conversations \citep{fraser_spoken_2018}. Language generation systems that use corpora and methods without awareness of emotions may generate callous, generic or even biased responses \cite{bender_stochastic_parrots_2021, sheng-etal-2019-woman}. Depending on the use case or type of system, it may be useful to stylistically vary responses, e.g., using shorter responses for spoken dialogue systems, longer responses if the system includes visual modality through a screen, or emotion-specific responses that appropriately address user sentiment.

Previous work on controlled text generation aimed at achieving stylistic goals has not focused on a schema-guided paradigm where specific content must be communicated correctly; instead, most work focuses more on unconstrained text-to-text generation without explicit meaning representations \cite{Liu2021DExpertsDC, Krause2020GeDiGD, Keskar2019CTRLAC, ghazvininejad-etal-2017-hafez}. Meanwhile, work on schema-guided NLG has primarily focused on generating fluent outputs that achieve low semantic error rather than achieving stylistic goals \cite{kale-rastogi-2020-template, Peng2020FewshotNL, du-etal-2020-schema}. In this paper, we hope to fill the gap on stylistic control and evaluation for  schema-guided NLG. Our contributions in this paper are three-fold: 
\begin{enumerate}
    \item We describe how we pre-process and annotate style parameters within the Schema-guided Dialogue (SGD) dataset \cite{rastogi2019towards}.
    \item We experiment with controlling different styles with various controlled text generation methods that aim to preserve fluency and semantic correctness.
    \item We present results with a broad range of evaluation methods, including a detailed human evaluation.
\end{enumerate}

Specifically, we consider three types of methods: \textit{conditional training}, \textit{guided fine-tuning}, and \textit{guided decoding}. We show that conditional training (CT) can be used for both very lexically-defined (e.g., point-of-view) and more semantically complex styles (e.g., empathy). However, CT introduces the need to re-train new models per style and is more effective at learning styles with strong lexical characteristics (e.g., specific language patterns or vocabulary). For guided fine-tuning, we explore the Plug-and-Play Language Model (PPLM) \cite{Dathathri2020PlugAP}, but show that it requires careful hyper-parameter turning and is prone to degeneration. For guided decoding, we evaluate the beam search weighted decoding (BSWD) method and show that it performs best overall on measures of style accuracy for semantically complex styles. The results suggest that unlike style control for unconstrained text generation where no specific content needs to be communicated, style control under the schema-guided paradigm has stronger restrictions on the degree of freedom allowed for content generation. We show that methods that disentangle content generation and style variations, especially for more semantically complex styles, result in better overall performance on semantic and stylistic control. 
	\section{Related Work}

Controllable text generation is an emerging research field. Current methods for controlling styles in text generation involve learning a conditional generative model or designing an appropriate decoding strategy. There are many methods proposed for learning a good conditional generative model. These include conditional training \citep{kikuchi-etal-2016-controlling, ficler_controlling_2017, Keskar2019CTRLAC, see_what_2019}, fine-tuning language models with external attribute models or side models \citep{Dathathri2020PlugAP, Zhang2020SideTuningAB}, fine-tuning models with reinforcement learning and human feedback \citep{Ziegler2019FineTuningLM}, training generative adversarial models \citep{Yu2017SeqGANSG}, and training variational auto-encoders \citep{Yu2017SeqGANSG, Hu2017TowardCG}. 

Others have worked on designing a good decoding strategy to guide generation, where the decoding procedure is guided by a scoring function or discriminator. These include weighted decoding \citep{ghazvininejad-etal-2017-hafez, Holtzman2018LearningTW, see_what_2019} and guided generation \citep{Krause2020GeDiGD}. Other lines of work include curating training data with rich style markup to facilitate training models with explicit stylistic supervision \citet{oraby_curate_2019, oraby_controlling_2018}. While this previous work does focus on controllable text generation, most work has been carried out in a text-to-text generation setting, without specific semantic constraints. Instead, we focus on the task-oriented dialogue framework where specific values must be communicated, and conduct a rigorous evaluation of different methods and their efficacy on different forms of style generation. Other recent work has explored adding additional information such as chit-chat data to task-oriented dialogue \cite{sun-etal-2021-adding, Madotto2020AttentionOP} and could potentially provide new opportunities for stylistic control.
	\section{Data Collection and Annotation}
\label{sec:data}
\subsection{Schema-to-Template pairs}
We use the Schema-guided Dialogue (SGD) dataset\footnote{The Schema-guided Dialogue Dataset: \url{https://github.com/google-research-datasets/dstc8-schema-guided-dialogue}} to create a rich corpus of schema-to-template pairs. This dataset is one of the largest publicly available corpora of annotated multi-domain, task-oriented dialogues \citep{rastogi2019towards}. Each dialogue in the data is represented as a list of user and system utterances. We use only the system-side utterances and annotations since we are focused on system-side generation.

Table \ref{table:sgd-data-sample} shows an example pre-processed data instance and the final flattened input. To create schema-to-template pairs, we follow the pre-processing steps outlined in \citet{du-etal-2020-schema} with the following modifications: (1) we replace the slot values with generic slot values without any slot type by adding a \texttt{\$} prefix and appending a increasing index (\textit{e.g.,} \texttt{San Jose} $\rightarrow$ \texttt{\$slot1}) for better generalization; and (2) we use only \textit{domain}, \textit{meaning representations (MRs)}, and \textit{slot description} as input data. Domain provides the context of the conversation (\textit{e.g.,} \texttt{Restaurants}); an MR contains a dialog act, a slot and a value (\textit{e.g.,} \texttt{OFFER(city=\$slot2)}); and the slot description describes the meaning of the slot in natural language. Table \ref{table:sgd-data-stats} in Appendix \ref{append:data-stats} summarizes the full statistics for the final pre-processed SGD dataset. In summary, we have 1,698 MRs and 118,715 example templates in the training set and 1,137 MRs and 34,598 templates in the test set.

\begin{table*}[!h]
	\footnotesize
	\centering
	\resizebox{.85\textwidth}{!}{
		\begin{tabularx}{\textwidth}{ lX|lX }
			\toprule
			\multicolumn{2}{c}{Original schema} & \multicolumn{2}{c}{Pre-processed schema} \\
			\midrule
			Service & \texttt{Restaurants\_1} & Domain & \texttt{Restaurants} \\
			Act & \texttt{OFFER} & Act & \texttt{OFFER}\\
			Slot & \texttt{restaurant\_name} & Slot & \texttt{restaurant\_name} \\
			Slot Description & \texttt{Name of the restaurant} & Slot Description & \texttt{Name of the restaurant} \\
			Values & \texttt{71 Saint Peter} & Value & \texttt{\$slot1} \\
			Act & \texttt{OFFER} & Act & \texttt{OFFER} \\
			Slot & \texttt{City} & Slot & \texttt{City} \\
			Slot Description & \texttt{City where the restaurant is located} & Slot Description & \texttt{City where the restaurant is located}\\
			Values & \texttt{San Jose} & Value & \texttt{\$slot2} \\
			\midrule
			Utterance &  \texttt{I see that at 71 Saint Peter there is a good restaurant which is in San Jose.} & Template &  \texttt{I see that at \$slot1 there is a good restaurant which is in \$slot2.} \\
			\midrule
		\end{tabularx}
	}
	
	\resizebox{.85\textwidth}{!}{
		\begin{tabulary}{\textwidth}{ LL }
			Flattened input after pre-processing & \texttt{restaurants offer restaurant\_name name of the restaurant \$slot1 offer city city where the restaurant is located \$slot2} \\
			\bottomrule
	\end{tabulary}}
	\caption{\label{table:sgd-data-sample} Sample system-side schema and the flat natural language strings.}
\end{table*}

For a single example, there are usually multiple dialogue acts as shown in Table \ref{table:sgd-data-sample}. An MR in the original data may also contain multiple values. In such cases, we flatten these MRs into multiple parts, each containing only one dialogue act. For instance, an MR that contains two values originally, \textit{e.g.,} \texttt{REQUEST(cuisine=[Mexican, Italian])} becomes two separate dialogue acts, \textit{e.g.,}  \texttt{REQUEST(cuisine=\$slot1)}, \texttt{REQUEST(cuisine=\$slot2)}. Templates are obtained by delexicalizing the utterances with generic slot values (\textit{e.g.,} \texttt{\$slot1 is a good restaurant}). Finally, we flatten the input data into flat natural language strings similar to \citet{budzianowski-vulic-2019-hello}.

\subsection{Style Parameters}
In order to perform style control, we also need to annotate style parameters for the SGD dataset. Style parameters are features of text that are stylistically expressive. These parameters can be roughly identified at \textit{\bfseries lexical} (vocabulary and words), \textit{\bfseries syntactic} (sentence structure) and \textit{\bfseries semantic} (abstract meaning/emotion) levels \citep{verma_lexical_2019}. We focus primarily on \textit{\bfseries lexical} and \textit{\bfseries semantic} features. Specifically, we characterize lexical style parameters as low-level linguistic features that can be derived from the text directly such as \textit{word count} and \textit{number of adjectives}, and semantic style parameters as high-level styles such as \textit{sentiment} that are more complex to characterize. 

\paragraph{Lexical style parameters}
Table \ref{table:lexical-styles} summarizes the lexical styles we annotate for the SGD data and the description of each parameter. In total, we automatically annotate six lexical style parameters for the SGD data. Similar to \citet{zhang_learning_2018}, the parameter ``\textit{has rare word}'' uses the maximum Normalized Inverse Document Frequency (NIDF) to determine whether or not a template contains words that are used less frequently in the corpus.\footnote{Appendix \ref{append:data-stats} includes the details of the NIDF calculation.} The complete data distribution for all the style parameters is included in Table \ref{table:data-distribution} in Appendix \ref{append:data-stats}.

\begin{table*}[!h]
    \footnotesize
    \centering
    \resizebox{.85\textwidth}{!}{
    \begin{tabularx}{\textwidth}{ lXl }
        \toprule
        \textbf{Style Parameter} & \textbf{Description} & \textbf{Condition} \\
        \midrule
            Short & Number of tokens in the template. The condition is based roughly on 25th percentile of the number of tokens per template in the training data. & $\le$ 7 \\
        \midrule
            Long & Number of tokens in the template. The condition is based roughly on 75th percentile of the number of tokens per template in the training data. & $\ge$ 15 \\
        \midrule
            Has Rare Word & Whether or not the template contains very specific or rarely used words compared to other templates in the corpus & Max NIDF $\ge$ 0.5 \\
        \midrule
            First Person Pronouns & Whether or not the template contains first person pronouns: I, me, my, mine & - \\
        \midrule
            Second Person Pronouns & Whether or not the template contains second person pronouns: you, your, yours & - \\
        \midrule
            Descriptive & Number of adjectives in the template & $>$ 2 \\
        \bottomrule
    \end{tabularx}
    }
    \caption{\label{table:lexical-styles} Lexical style parameters and possible values.}
\end{table*}

\paragraph{Semantic style parameters}
Unlike lexical parameters which consider explicit features such as vocabulary, semantic parameters of style are less lexically-defined. As a result, it is generally harder to annotate these parameters directly from the original data without auxiliary information. In this work, we consider the following semantic parameters: \textit{\bfseries formality}, \textit{\bfseries negative sentiment}, \textit{\bfseries positive sentiment} and \textit{\bfseries empathy}. Formality and sentiment are common stylistic parameters studied in the stylistic control NLG literature. We also include empathy as an interesting and complex style studied in recent work \cite{9362067, majumder-etal-2020-mime, lin_moel_2019, zhou-wang-2018-mojitalk}. We train a classifier for each of the four styles and annotate the utterances in SGD with these features. We include more details about the classifiers in Section \ref{sec:baseline-model} and show additional information about the dataset used to train each classifier in Table \ref{table:semantic-styles} of Appendix \ref{append:data-stats}.
	\section{Baseline Model}
\label{sec:baseline-model}

\paragraph{Language model}
Given a sequence of tokens $x = \big\{x_1, \cdots, x_t \big\}$, the goal of the language model is to model the joint probability of the sequence $p(x) = p(x_1, \cdots, x_t)$. The joint probability $p(x)$ is often factorized in terms of the product of conditional probabilities using the chain rule of probability $P(x) = \prod_{t=1}^T P(x_T|x_0, \cdots, x_{T-1})$ \cite{NIPS2000_728f206c}.
In recent years, transformer-based models have been used widely to model these conditional probabilities \citep{NIPS2017_3f5ee243}. In this work, we use GPT-2\footnote{GPT-2 small from HuggingFace: \url{https://huggingface.co/transformers/model_doc/gpt2.html}} as our baseline language model \citep{radford_language_nodate} and fine-tune the GPT-2 model with the processed SGD data using a flat representation with the beginning of sequence, separator, and end of sequence special tokens.\footnote{\textit{e.g.,} ``\texttt{[BOS]} flattened-schema-tokens \texttt{[SEP]} template-tokens \texttt{[EOS]}''}

\paragraph{Semantic style classifiers}
The classifiers used to annotate semantic parameters are single layer classifiers. To train each classifier, we encode the input $x = \big\{x_1, \cdots, x_t\big\}$ of length $t$ using the baseline GPT-2 model described above and obtain the last hidden layer $o_t$ for all time steps $t$. We then take the average representation across time, denoted $\bar{o}_t$, and train the classifier to predict the target label (\textit{e.g.,} formal vs. informal) from the average representation: $f(\bar{o}_t) = f\Big( \sum_{t=1}^T \frac{o_t}{T} \Big)$. The classifiers are used for annotating the semantic parameters of the SGD data and for the style control models described in Section \ref{sec:methods}.
	\section{Style Controlled Text Generation}
\label{sec:methods}
In this work, we require a controlled generation method that is able to simultaneously render the semantic content given the schemata while achieving the desired stylistic goals. We also require the method to be both stable (preserve the fluency of the response even when the stylistic goals are not met) and general-purpose (can be applied to many styles). Under these requirements and constraints, we discuss three types of controlled generation methods to achieve these goals: \emph{conditional training}, \emph{guided fine-tuning}, and \emph{guided decoding}, and compare their performance in Section \ref{sec:experiments}. To our knowledge, our work is the first to systematically study the effectiveness of these control methods for schema-guided NLG.

\subsection{Conditional Training (CT)}
Controllable generation entails modeling $p(x|a)$, where $a$ is a control variable and $x$ is the generated sequence. However, a pre-trained language model such as GPT-2 is only trained to learn $p(x)$. On the other hand, conditional training \citep{kikuchi-etal-2016-controlling, peng_towards_2018, fan_controllable_2018, see_what_2019} refers to directly learning the conditional generative model $p(x|a)$. The results are of high quality because the model is trained to directly maximize $p(x|a)$, but this comes at the expense of fixing the control variable upfront and of re-training the entire model for each new control variable.

To perform style control, we fine-tune the baseline GPT-2 with the conditional training method. Specifically, each input in the training set is annotated with the variable $a$ that we wish to control, \emph{e.g.,} the length (\emph{short, long}) of the input. The value of the control variable $a$ is then added to model vocabulary as a special token (\emph{e.g.,} \verb|[LENGTH_SHORT]|) and appended to the meaning representation after the \verb|[BOS]| special token. The model then learns an embedding for each value of $a$ and learns to generate $x = \big\{x_1, \cdots, x_t \big\}$ conditioned on a value of $a$ and the given meaning representation by optimizing cross-entropy loss.

%%%%%%%%%%%%%%%%%%%%%%%%%%%%%%%%%
\subsection{Guided Fine-tuning}
\label{subsec:guided-fine-tune}
Unlike conditional training that requires fine-tuning an entire GPT-2 model per style, guided fine-turning refers to methods that require only fine-tuning a smaller set of the parameters while the majority of the base model stays fixed. In this paper, we consider the recent Plug-and-Play Language Model (PPLM) \citep{Dathathri2020PlugAP}. In guided fine-tuning methods, the conditional probability $p(x|a) \propto p(x)p(a|x)$ is obtained by fine-tuning the base language model (LM) using an auxiliary discriminator that explicitly models $p(a|x)$. In our work, we use the semantic style classifiers described in Section \ref{sec:baseline-model} for the discriminator $p(a|x)$ and the GPT-2 model for the base LM. 

The major difficulty of the PPLM method is the problem of degeneration -- output that is ungrammatical, incoherent, or repetitive. In practice, we observe that PPLM is prone to generating ungrammatical outputs or getting stuck in a repetitive loop if the hyper-parameters are not carefully tuned. We illustrate the effect of hyper-parameters tuning and the degeneration problem in Appendix \ref{append:hyper-parameters}.

%%%%%%%%%%%%%%%%%%%%%%%%%%%%%%%%%
\subsection{Guided Decoding}
While conditional training and guided fine-tuning require fine-tuning the base language model, weighted decoding is applied only at decoding time, requiring no change to the base language model. To control the generation, it re-ranks the probability of words based on a scoring function or discriminator.

\paragraph{Weighted Decoding (WD)}
In weighted decoding \citep{ghazvininejad-etal-2017-hafez}, at  time step $t$, the distribution of the next token $x_{t+1}$ is re-weighted by a semantic style classifier that models $p(a|x)$. The probability of each possible next word $w$ in the vocabulary given the control variable $a$ is then re-computed as
\begin{align}
    p(w|a) = \text{Softmax} \Big(p(w) \, p(a|w) \Big). \label{eq:wd}
\end{align}
Here $p(w)$ is the probability of the word $w$ calculated by the base language model as the next token given the generated sequence $\big\{x_1, \cdots, x_t \big\}$, and $p(a|w)$ is the probability of the word $w$ associated with the control variable $a$. 

\paragraph{Beam search weighted decoding (BSWD)}
The weighted decoding method described above takes the highest scoring item at each time step. While this approach is effective, it is often non-optimal and can limit the diversity of the generated text. To mitigate this limitation, we can increase the search space at generation time using the beam search algorithm. Given a fixed beam width parameter $B$, the beam search algorithm selects $B$ best alternatives with the highest probability for an input sequence at each time step. Therefore, the original weighted decoding approach described above is a special case of the beam search algorithm with $B=1$.

Finally, we note that in Eq. \ref{eq:wd}, the style classifier is only conditioned on the next possible token $w$ but not the entire past sequence, \emph{i.e.,} the next possible token $w$ plus the text that has been generated $\big\{x_1, \cdots, x_t, w \big\}$. Empirically, in both WD and BSWD, we observe that maximizing the probability of the desired style by greedily considering only the next generated token, rather than the entire sequence of previously generated tokens, yielded better performance on the SGD data. When the entire sequence representation is used, we find that the re-weighting of the distribution is usually not strong enough to successfully match the desired stylistic goal.
	\section{Experiments}
\label{sec:experiments}
In this section, we show the experimental results of the methods described in Section \ref{sec:methods} for controlling the styles described in Section \ref{sec:data} on the SGD data. The baseline GPT-2 model is fine-tuned on the training set with no control variables, and the conditional training model is fine-tuned with control variable special tokens, \emph{e.g.,} \verb|LENGTH_SHORT|. Our evaluation is tested on the test set of 1,137 MRs. We focus on controlling a single style at a time in this experiment; however, it is also possible to control for multiple styles -- we include details on multiple-style control experiments in Appendix \ref{append:additional-experiments} (with sample outputs in Appendix Table \ref{table:ct-bswd-example}).

\subsection{Automatic Evaluation Metrics}
We focus on evaluating three key dimensions: style accuracy, fluency, and semantic correctness.

\paragraph{Style accuracy}
To evaluate how effective each controlled generation method is per style, we use the {\it style accuracy} metric, or the percentage of outputs that conform to the required input style. For lexical styles, this is simply computed using the conditions in Table \ref{table:lexical-styles}. For semantic styles, we classify the generated text using the corresponding style classifier and check if the predicted style matches the desired style value. For instance, if the predicted sentiment for generated text with the ``positive sentiment'' control code does not match the ``positive'' label, then it is considered incorrect.

\paragraph{Response fluency}
We use {\it BLEU} score (n-gram precision with brevity penalty) \citep{papineni-etal-2002-bleu} as a measurement of the response fluency. We acknowledge that lexical overlap metrics are poor measures of quality \citep{novikova-etal-2017-need}; however, we include BLEU for completeness and further evaluate quality through human judgments. 

\paragraph{Semantic correctness}
We use {\it slot error rate (SER)} \citep{luong-etal-2015-effective} to measure the semantic correctness of the generated response as compared to the given MR. SER measures the ratio of semantic errors that the model makes by finding the total number of slot mistakes (deletions, repetitions, and hallucinations) in the generated text (lower is better). SER here only considers slots that have explicit values that must be realized (e.g., {\tt\$slotN}).

\subsection{Lexical Style Automatic Evaluation}
\label{subsec:lexical}
We evaluate lexical styles with only conditional training as the rest of the methods include semantic style classifiers and are thus not applicable. Table \ref{table:lexical-style-eval} summarizes the performance of conditional training for the six lexical styles. The style accuracy for most styles is generally high, between 80\% to nearly 100\%, especially for styles marked explicitly by specific words, such as first and second person pronouns. However, we observe that ``\textit{descriptive}'' has a particularly low accuracy. First, the majority of the references in the training data (95\%) have less than two adjectives, making it difficult for the model to learn this kind of style effectively.\footnote{The full data distribution can be found in Appendix \ref{append:data-stats} Table \ref{table:data-distribution}.} Secondly, we observe that conditional training is particularly effective when the style exhibits a clear syntactic characteristic (\textit{e.g.}, length) or a particular set of vocabulary (\textit{e.g.}, pronouns); however, this is not the case for the ``\textit{descriptive}'' style. 

The fluency of the generated text with style control drops slightly as compared to no style control. Having style control often makes the generated text different from its matching responses, \textit{i.e.}, adding extra content or changing linguistic characteristics. Since BLEU score only considers lexical overlap, this behavior is expected. Finally, we see that there is not much of a performance drop in semantic correctness with respect to SER. The experimental results show that for lexical styles with a clear syntactic pattern or vocabulary, CT can be quite effective.

Table \ref{table:ct-example} illustrates example outputs using CT when controlling for ``\textit{short}'', ``\textit{long}'' and ``\textit{has rare word}'' styles.\footnote{Example outputs for other lexical styles are included in Appendix \ref{append:additional-experiments} Table \ref{table:additional-ct-outputs}.} Interestingly, we see that when asked for a longer response, the model starts to hallucinate extra content (but not slots) not given in the MR in order to satisfy the control variable. This also translates to slightly lower BLEU scores and a higher SER. Methods to enforce better implicit constraints to increase fluency and semantic correctness are important points for future work.

\begin{table}[!h]
    \centering
    \scriptsize
    \begin{tabulary}{.48\textwidth}{ LCCC }
        \toprule
        \textbf{Style} & \textbf{Style Acc.} & \textbf{BLEU $\uparrow$} & \textbf{SER $\downarrow$} \\
        \midrule
        Baseline (no style control) & - & 0.480 & 0.010 \\
        \midrule
        Short & 96.4\% & 0.205 & 0.009 \\
        \midrule
        Long & 80.1\% & 0.381 & 0.012 \\
        \midrule
        Has Rare Word & 89.5\% & 0.377 & 0.010 \\
        \midrule
        First Person Pronouns & 99.9\% & 0.365 & 0.011 \\
        \midrule
        Second Person Pronouns & 99.9\% & 0.447 & 0.011 \\
        \midrule
        Descriptive & 19.6\% & 0.378 & 0.012 \\
        \bottomrule
    \end{tabulary}
    \caption{\label{table:lexical-style-eval} Evaluation of lexical style parameters with conditional training.}
\end{table}

\begin{table*}[!h]
    \centering
    \resizebox{.9\textwidth}{!}{
    \begin{tabulary}{\textwidth}{ ll }
        \toprule
        \multicolumn{2}{l}{\textbf{MR:} \texttt{OFFER(restaurant\_name=\$slot1), OFFER(city=\$slot2)}} \\
        \midrule
        W/o Style Control & \verb|$slot1| is a nice restaurant in \verb|$slot2| that serves curry. \\
        Short & \verb|$slot1| is a nice restaurant in \verb|$slot2|. \\
        Long & Okay! the restaurant, \verb|$slot1| located in \verb|$slot2| is a good one and serves Taiwanese dishes. \\
        \midrule
        
        \multicolumn{2}{l}{\textbf{MR:} \texttt{OFFER(address=\$slot1), OFFER(rating=\$slot2)}} \\
        \midrule
        W/o Style Control &  There is a nice house at \verb|$slot1| with a \verb|$slot2| rating. \\
        Has Rare Word & There is a lovely residence located at \verb|$slot1|. It has a rating of \verb|$slot2|. \\
        \bottomrule
    \end{tabulary}}
    \caption{\label{table:ct-example} Example outputs for conditional training with lexical styles ``\textit{short}'', ``\textit{long}'' and ``\textit{has rare word}'' (more content hallucinations for ``\textit{long}''). Example outputs for other lexical styles are included in Appendix \ref{append:additional-experiments} Table \ref{table:additional-ct-outputs}.}
\end{table*}

\subsection{Semantic Style Automatic Evaluation}
Table \ref{table:semantic-style-eval} summarizes the main results for semantic style evaluation using CT, PPLM, and BSWD. Since CT is trained to directly maximize conditional probability, it frequently has a higher BLEU score and a lower SER across different styles with the exception of ``\textit{formal}'' BLEU. We note that formality is rather lexically-defined, exhibiting characteristic keywords such as ``\textbf{please}'', which are frequently picked up by the model, resulting in a particularly high style accuracy for ``\textit{formal}'' responses and a lower BLEU score.

For the three more semantically complex styles, ``\textit{positive sentiment}'', ``\textit{negative sentiment}'' and ``\textit{empathy}'', we see that BSWD achieves a higher style accuracy than CT (at the cost of a lower BLEU and slightly higher semantic error). We note, however, that the drop in BLEU is expected: as the output is steered towards a certain style, its n-gram overlap with the references is more likely to decrease -- thus, we use these automatic metrics as an evaluation guide, but leave a more rigorous evaluation to human judgment in the next section. Finally, with careful hyper-parameter tuning, PPLM can achieve similar performance to BSWD on BLEU and SER but does worse on style accuracy. Increasing the style accuracy for PPLM worsens the BLEU and SER score significantly and thus we do not consider it in our human evaluation. In summary, our automatic evaluation shows that for semantic styles, BSWD gives us a good trade-off between consistent style accuracy and semantic fidelity across styles.

Table \ref{table:semantic-example} illustrates example outputs for the three semantic styles using BSWD (with additional examples including combining multiple styles in Appendix \ref{append:multiple-styles}). In general, styles that encapsulate complex phenomena such as ``\textit{empathy}'' are harder to generate as shown by their lower style accuracy; nevertheless, we are able to preserve fluency and semantic correctness in most cases.

\begin{table}[!h]
    \centering
    \scriptsize
    \begin{tabulary}{.48\textwidth}{ LLCCC }
        \toprule
         \textbf{Style} & \textbf{Model} & \textbf{Style Acc.} & \textbf{BLEU $\uparrow$} & \textbf{SER $\downarrow$} \\
        
        \midrule
        Baseline & - & - & 0.480 & 0.010 \\

        \midrule
         \multirow{3}{*}{Formal} & CT & \textbf{70.0\%} & 0.461 & \textbf{0.009} \\
         & PPLM & 17.5\% & 0.390 & 0.016 \\
         & BSWD & 42.5\% & \textbf{0.496} & 0.021 \\
         
        \midrule
         \multirow{3}{*}{Negative} & CT & 7.5\% & \textbf{0.409} & \textbf{0.012} \\
         & PPLM & 2.5\% & 0.373 & 0.017 \\
         & BSWD & \textbf{25.0\%} & 0.359 & 0.021 \\
         \midrule
         
         \multirow{3}{*}{Positive} & CT & 17.5\% & \textbf{0.442} & \textbf{0.009} \\
         & PPLM & 32.5\% & 0.371 & 0.026 \\
         & BSWD & \textbf{65.0\%} & 0.369 & 0.021 \\
         
         \midrule
         \multirow{3}{*}{Empathy} & CT & 21.5\% & \textbf{0.449} & \textbf{0.009} \\
         & PPLM & 11.0\% & 0.368 & 0.021 \\
         & BSWD & \textbf{31.0\%} & 0.394 & 0.017 \\
         
        \bottomrule
    \end{tabulary}
    \caption{\label{table:semantic-style-eval} Evaluation of semantic styles.}
\end{table} 

\begin{table*}[h]
    \centering
    \resizebox{.7\textwidth}{!}{
    \begin{tabulary}{\textwidth}{ ll }
        \toprule
        \multicolumn{2}{l}{{\bf MR:} \verb|REQUEST(where_to=none)|} \\
        \midrule
        W/o style control & What city are you staying? \\
        Formal &  What city are you planning to stay in? \\
        \midrule
        
        \multicolumn{2}{l}{{\bf MR:} \verb|OFFER(address=$slot1), OFFER(rating=$slot2)|} \\
        \midrule
        W/o style control & There is a house at \verb|$slot1| with a rating of \verb|$slot2|. \\ 
        Positive & There is a nice house with a \verb|$slot2| rating located at \verb|$slot1|. \\
        \midrule
        
        \multicolumn{2}{l}{{\bf MR:} \verb|INFORM(rating=$slot1), NOTIFY_FAILURE(null=none)|} \\
        \midrule
        W/o style control & The rating is \verb|$slot1|. I was unable to make a reservation. \\
        Empathy & The rating is \verb|$slot1|. I'm sorry, but I couldn't make the reservation. \\
        \bottomrule
    \end{tabulary}}
    \caption{\label{table:semantic-example} Example outputs of semantic styles using beam search weighted decoding (BSWD).}
\end{table*}

\subsection{Human Evaluation}
\label{subsec:human-eval}
We focus on human evaluation for our semantic styles since they are the most inherently subjective.\footnote{More details on the human evaluation design are in Appendix \ref{append:human-eval}.} We pick a subset of our semantic styles to evaluate, specifically, \textit{formal}, \textit{negative} and \textit{positive}. We also focus on the evaluation of CT and BSWD only since they have an overall better performance in the automatic evaluation and are simpler in nature (\textit{e.g.,} less hyper-parameter tuning). To evaluate style, we ask three human annotators to rate:
\begin{itemize}
    \item \textbf{Style Rating (Sty. Rat.)}: How closely the response matches the given style (1 being not at all, 5 being very closely).
    \item \textbf{Fluency (Flu.)}: The fluency of the generated response (1 being low, 5 being high).
    \item \textbf{Semantic Error}: Slot errors in the generated response using a 0/1 scale. For analysis purposes, we further break down the items marked ``0'' into four error types to understand the strengths and weaknesses of each method. For each type of error, the output is only marked ``1'' if there are no issues.
    \begin{itemize}
        \item \textbf{Deletion (Del.)}: Whether the response drops any slot values from the MR.
        \item \textbf{Repetition (Rep.)}: Whether the response repeats any slot values from the MR.
        \item \textbf{Content Hallucination (Cont. Hal.)}: Whether the response includes extra content not given in the MR.
        \item \textbf{Incorrect Slot Values (Inc. Slot)}: Whether the response includes any slot values not given in the MR (a specific type of hallucination).
    \end{itemize}
\end{itemize} 

Table \ref{table:human-eval} shows the aggregated evaluation (40 samples per style, with three judgments per sample). The results show that outputs generated by BSWD have a higher or comparable style rating compared to CT, confirming its ability to control semantic styles. We also observe that outputs from BSWD have a higher fluency score across all styles. This means that even though BSWD showed lower BLEU scores in automatic evaluation, its outputs are considered to be more natural and fluent in human evaluation. Finally, we see an interesting difference in error types when comparing CT and BSWD. In general, BSWD is more prone to deleting and hallucinating slots, while CT more frequently generates incorrect slot values. Since BSWD requires no change to the base language model, it is able to obtain a lower incorrect slot value error rate as compared to CT, which requires re-training the language model with a control code. On the other hand, it has a higher deletion and hallucination error rate since during decoding time, BSWD is free to insert or drop content in order to achieve the desired style.

\begin{table}[!h]
    \centering
    \scriptsize
    \begin{tabulary}{.48\textwidth}{ LLLLLLLL }
        \toprule
        & & & & \multicolumn{4}{c}{\textbf{Semantic Error}} \\
        \cmidrule(lr){5-8}
        \textbf{Style} & \textbf{Model} & \textbf{Sty. Rat.} & \textbf{Flu.} & \textbf{Del.} & \textbf{Rep.} & \textbf{Cont. Hal.} & \textbf{Inc. Slot} \\
         
         \midrule
         \multirow{2}{*}{Formal} & CT & 3.88 & 3.83 & \textbf{15\%} & 0\% & \textbf{13\%} & 3.75\% \\
         \cmidrule(lr){2-8}
         & BSWD & \textbf{3.95} & \textbf{4.30} & 19\% & 0\% & 15\% & \textbf{1.25\% }\\

         \midrule
         \multirow{2}{*}{Negative} & CT & 2.78 & 3.54 & 20\% & 0\% & 12.5\% & \textbf{2.5\%} \\
         \cmidrule(lr){2-8}
         & BSWD & \textbf{2.86} & \textbf{3.93} & 20\% & 0\% & 12.5\% & 5\% \\
         
         \midrule
         \multirow{2}{*}{Positive} & CT & \textbf{3.48} & 3.64 & \textbf{22.5\%} & 0\% & \textbf{22.5\%} & 12.5\% \\
         \cmidrule(lr){2-8}
         & BSWD & 3.42 & \textbf{3.84} & 27.5\% & 0\% & 30\% & \textbf{7.5\%} \\
        \bottomrule
    \end{tabulary}
    \caption{\label{table:human-eval} Human evaluation results for selected semantic styles and methods.}
\end{table}

	\section{Conclusion}
In this work, we focus on stylistic control and evaluation of schema-guided NLG. We discuss three different types of methods for style controlled text generation: conditional training (CT), guided fine-tuning (PPLM), and guided decoding (BSWD). We present a rich set of evaluations to quantify each method's ability to achieve various styles while preserving language fluency and minimizing slot errors. Our analysis shows that, in general, styles that encapsulate abstract ideas are naturally harder to generate (\textit{e.g.,} empathy), and methods that require careful hyper-parameters tuning may run into the problems of instability and degeneration (\textit{e.g.,} PPLM) while under-performing in style accuracy. The automatic and human evaluations suggest that simultaneously achieving stylistic goals and realizing schema information requires methods that allow us to separate content generation and stylistic variations. We show that CT and BSWD overcome some of these challenges and are effective at controlling several styles while maintaining good fluency and semantic correctness in most cases. CT is effective for lexical styles with strong syntactic characteristics or distinctive vocabulary sets, while BSWD excels at semantic styles that are more complex to characterize.

For future work, we are interested in extending our analysis to a larger number of styles, and in exploring techniques for understanding and representing styles that are more abstract, such as ``empathy'', as well as methods for generating those styles under a task-oriented dialogue framework. Additionally, since abstract styles may be characterized by multiple features (\textit{e.g.,} a combination of sentiment and descriptiveness), we are interested in studying how these underlying features can be represented and incorporated more accurately to improve overall semantic and stylistic control.
	\section*{Acknowledgments}
	We would like to thank Sofia Scharfenberg, Jasmin Rehm, and the rest of the Alexa Data Services team for all of their help with preparing and performing the human evaluation study.
	
	\bibliographystyle{acl_natbib}
	\bibliography{reference}

\begin{thebibliography}{48}
\expandafter\ifx\csname natexlab\endcsname\relax\def\natexlab#1{#1}\fi

\bibitem[{Bapna et~al.(2017)Bapna, Tur, Hakkani-Tur, and Heck}]{46223}
Ankur Bapna, Gokhan Tur, Dilek Hakkani-Tur, and Larry Heck. 2017.
\newblock Towards zero shot frame semantic parsing for domain scaling.
\newblock In \emph{Interspeech 2017}.

\bibitem[{Bender et~al.(2021)Bender, Gebru, McMillan-Major, and
  Shmitchell}]{bender_stochastic_parrots_2021}
Emily~M. Bender, Timnit Gebru, Angelina McMillan-Major, and Shmargaret
  Shmitchell. 2021.
\newblock \href {https://doi.org/10.1145/3442188.3445922} {On the dangers of
  stochastic parrots: Can language models be too big?}
\newblock In \emph{Proceedings of the 2021 ACM Conference on Fairness,
  Accountability, and Transparency}, FAccT '21, page 610–623, New York, NY,
  USA. Association for Computing Machinery.

\bibitem[{Bengio et~al.(2001)Bengio, Ducharme, and Vincent}]{NIPS2000_728f206c}
Yoshua Bengio, R\'{e}jean Ducharme, and Pascal Vincent. 2001.
\newblock \href
  {https://proceedings.neurips.cc/paper/2000/file/728f206c2a01bf572b5940d7d9a8fa4c-Paper.pdf}
  {A neural probabilistic language model}.
\newblock In \emph{Advances in Neural Information Processing Systems},
  volume~13. MIT Press.

\bibitem[{Budzianowski and Vuli{\'c}(2019)}]{budzianowski-vulic-2019-hello}
Pawe{\l} Budzianowski and Ivan Vuli{\'c}. 2019.
\newblock \href {https://doi.org/10.18653/v1/D19-5602} {Hello, it{'}s {GPT}-2 -
  how can {I} help you? towards the use of pretrained language models for
  task-oriented dialogue systems}.
\newblock In \emph{Proceedings of the 3rd Workshop on Neural Generation and
  Translation}, pages 15--22, Hong Kong. Association for Computational
  Linguistics.

\bibitem[{Buechel et~al.(2018)Buechel, Buffone, Slaff, Ungar, and
  Sedoc}]{Buechel18emnlp}
Sven Buechel, Anneke Buffone, Barry Slaff, Lyle Ungar, and Jo{\~{a}}o Sedoc.
  2018.
\newblock Modeling empathy and distress in reaction to news stories.
\newblock In \emph{Proceedings of the 2018 Conference on Empirical Methods in
  Natural Language Processing (EMNLP 2018)}.

\bibitem[{Colin et~al.(2016)Colin, Gardent, M{'}rabet, Narayan, and
  Perez-Beltrachini}]{colin-etal-2016-webnlg}
Emilie Colin, Claire Gardent, Yassine M{'}rabet, Shashi Narayan, and Laura
  Perez-Beltrachini. 2016.
\newblock \href {https://doi.org/10.18653/v1/W16-6626} {The {W}eb{NLG}
  challenge: Generating text from {DBP}edia data}.
\newblock In \emph{Proceedings of the 9th International Natural Language
  Generation conference}, pages 163--167, Edinburgh, UK. Association for
  Computational Linguistics.

\bibitem[{Dathathri et~al.(2020)Dathathri, Madotto, Lan, Hung, Frank, Molino,
  Yosinski, and Liu}]{Dathathri2020PlugAP}
Sumanth Dathathri, Andrea Madotto, Janice Lan, Jane Hung, Eric Frank, Piero
  Molino, Jason Yosinski, and Rosanne Liu. 2020.
\newblock \href {https://openreview.net/forum?id=H1edEyBKDS} {Plug and play
  language models: A simple approach to controlled text generation}.
\newblock In \emph{International Conference on Learning Representations}.

\bibitem[{Du et~al.(2020)Du, Oraby, Perera, Shen, Narayan-Chen, Chung,
  Venkatesh, and Hakkani-Tur}]{du-etal-2020-schema}
Yuheng Du, Shereen Oraby, Vittorio Perera, Minmin Shen, Anjali Narayan-Chen,
  Tagyoung Chung, Anushree Venkatesh, and Dilek Hakkani-Tur. 2020.
\newblock \href {https://www.aclweb.org/anthology/2020.inlg-1.35}
  {Schema-guided natural language generation}.
\newblock In \emph{Proceedings of the 13th International Conference on Natural
  Language Generation}, pages 283--295, Dublin, Ireland. Association for
  Computational Linguistics.

\bibitem[{Du{\v{s}}ek et~al.(2018)Du{\v{s}}ek, Novikova, and
  Rieser}]{dusek-etal-2018-findings}
Ond{\v{r}}ej Du{\v{s}}ek, Jekaterina Novikova, and Verena Rieser. 2018.
\newblock \href {https://doi.org/10.18653/v1/W18-6539} {Findings of the {E}2{E}
  {NLG} challenge}.
\newblock In \emph{Proceedings of the 11th International Conference on Natural
  Language Generation}, pages 322--328, Tilburg University, The Netherlands.
  Association for Computational Linguistics.

\bibitem[{Fan et~al.(2018)Fan, Grangier, and Auli}]{fan_controllable_2018}
Angela Fan, David Grangier, and Michael Auli. 2018.
\newblock \href {https://doi.org/10.18653/v1/W18-2706} {Controllable
  {Abstractive} {Summarization}}.
\newblock In \emph{Proceedings of the 2nd {Workshop} on {Neural} {Machine}
  {Translation} and {Generation}}, pages 45--54, Melbourne, Australia.
  Association for Computational Linguistics.

\bibitem[{Ficler and Goldberg(2017)}]{ficler_controlling_2017}
Jessica Ficler and Yoav Goldberg. 2017.
\newblock \href {https://doi.org/10.18653/v1/W17-4912} {Controlling
  {Linguistic} {Style} {Aspects} in {Neural} {Language} {Generation}}.
\newblock In \emph{Proceedings of the {Workshop} on {Stylistic} {Variation}},
  pages 94--104, Copenhagen, Denmark. Association for Computational
  Linguistics.

\bibitem[{Fraser et~al.(2018)Fraser, Papaioannou, and
  Lemon}]{fraser_spoken_2018}
Jamie Fraser, Ioannis Papaioannou, and Oliver Lemon. 2018.
\newblock Spoken {Conversational} {AI} in {Video} {Games}: {Emotional}
  {Dialogue} {Management} {Increases} {User} {Engagement}.
\newblock In \emph{Proceedings of the 18th {International} {Conference} on
  {Intelligent} {Virtual} {Agents}}.

\bibitem[{Ghazvininejad et~al.(2017)Ghazvininejad, Shi, Priyadarshi, and
  Knight}]{ghazvininejad-etal-2017-hafez}
Marjan Ghazvininejad, Xing Shi, Jay Priyadarshi, and Kevin Knight. 2017.
\newblock \href {https://www.aclweb.org/anthology/P17-4008} {{H}afez: an
  interactive poetry generation system}.
\newblock In \emph{Proceedings of {ACL} 2017, System Demonstrations}, pages
  43--48, Vancouver, Canada. Association for Computational Linguistics.

\bibitem[{Holtzman et~al.(2018)Holtzman, Buys, Forbes, Bosselut, Golub, and
  Choi}]{Holtzman2018LearningTW}
Ari Holtzman, Jan Buys, Maxwell Forbes, Antoine Bosselut, D.~Golub, and Yejin
  Choi. 2018.
\newblock Learning to write with cooperative discriminators.
\newblock \emph{ArXiv}, abs/1805.06087.

\bibitem[{Hu et~al.(2017)Hu, Yang, Liang, Salakhutdinov, and
  Xing}]{Hu2017TowardCG}
Zhiting Hu, Zichao Yang, Xiaodan Liang, R.~Salakhutdinov, and E.~Xing. 2017.
\newblock Toward controlled generation of text.
\newblock In \emph{ICML}.

\bibitem[{Kale and Rastogi(2020)}]{kale-rastogi-2020-template}
Mihir Kale and Abhinav Rastogi. 2020.
\newblock \href {https://doi.org/10.18653/v1/2020.emnlp-main.527} {Template
  guided text generation for task-oriented dialogue}.
\newblock In \emph{Proceedings of the 2020 Conference on Empirical Methods in
  Natural Language Processing (EMNLP)}, pages 6505--6520, Online. Association
  for Computational Linguistics.

\bibitem[{Keskar et~al.(2019)Keskar, McCann, Varshney, Xiong, and
  Socher}]{Keskar2019CTRLAC}
Nitish~Shirish Keskar, Bryan McCann, Lav~R. Varshney, Caiming Xiong, and
  Richard Socher. 2019.
\newblock Ctrl: A conditional transformer language model for controllable
  generation.
\newblock \emph{ArXiv}, abs/1909.05858.

\bibitem[{Kikuchi et~al.(2016)Kikuchi, Neubig, Sasano, Takamura, and
  Okumura}]{kikuchi-etal-2016-controlling}
Yuta Kikuchi, Graham Neubig, Ryohei Sasano, Hiroya Takamura, and Manabu
  Okumura. 2016.
\newblock \href {https://doi.org/10.18653/v1/D16-1140} {Controlling output
  length in neural encoder-decoders}.
\newblock In \emph{Proceedings of the 2016 Conference on Empirical Methods in
  Natural Language Processing}, pages 1328--1338, Austin, Texas. Association
  for Computational Linguistics.

\bibitem[{Krause et~al.(2020)Krause, Gotmare, McCann, Keskar, Joty, Socher, and
  Rajani}]{Krause2020GeDiGD}
Ben Krause, Akhilesh~Deepak Gotmare, B.~McCann, N.~Keskar, Shafiq~R. Joty,
  R.~Socher, and Nazneen Rajani. 2020.
\newblock Gedi: Generative discriminator guided sequence generation.
\newblock \emph{ArXiv}, abs/2009.06367.

\bibitem[{Lin et~al.(2021)Lin, Liu, Moon, Crook, Zhou, Wang, Yu, Madotto, Cho,
  and Subba}]{lin-etal-2021-leveraging}
Zhaojiang Lin, Bing Liu, Seungwhan Moon, Paul Crook, Zhenpeng Zhou, Zhiguang
  Wang, Zhou Yu, Andrea Madotto, Eunjoon Cho, and Rajen Subba. 2021.
\newblock \href {https://doi.org/10.18653/v1/2021.naacl-main.448} {Leveraging
  slot descriptions for zero-shot cross-domain dialogue {S}tate{T}racking}.
\newblock In \emph{Proceedings of the 2021 Conference of the North American
  Chapter of the Association for Computational Linguistics: Human Language
  Technologies}, pages 5640--5648, Online. Association for Computational
  Linguistics.

\bibitem[{Lin et~al.(2019)Lin, Madotto, Shin, Xu, and Fung}]{lin_moel_2019}
Zhaojiang Lin, Andrea Madotto, Jamin Shin, Peng Xu, and Pascale Fung. 2019.
\newblock {MoEL}: {Mixture} of {Empathetic} {Listeners}.
\newblock In \emph{Proceedings of the 2019 {Conference} on {Empirical}
  {Methods} in {Natural} {Language} {Processing} and the 9th {International}
  {Joint} {Conference} on {Natural} {Language} {Processing}
  ({EMNLP}-{IJCNLP})}, pages 121--132, Hong Kong, China.

\bibitem[{Liu et~al.(2021)Liu, Sap, Lu, Swayamdipta, Bhagavatula, Smith, and
  Choi}]{Liu2021DExpertsDC}
Alisa Liu, Maarten Sap, Ximing Lu, Swabha Swayamdipta, Chandra Bhagavatula,
  Noah~A. Smith, and Yejin Choi. 2021.
\newblock Dexperts: Decoding-time controlled text generation with experts and
  anti-experts.

\bibitem[{Luong et~al.(2015)Luong, Pham, and
  Manning}]{luong-etal-2015-effective}
Thang Luong, Hieu Pham, and Christopher~D. Manning. 2015.
\newblock \href {https://doi.org/10.18653/v1/D15-1166} {Effective approaches to
  attention-based neural machine translation}.
\newblock In \emph{Proceedings of the 2015 Conference on Empirical Methods in
  Natural Language Processing}, pages 1412--1421, Lisbon, Portugal. Association
  for Computational Linguistics.

\bibitem[{Madotto et~al.(2020)Madotto, Lin, Wu, Shin, and
  Fung}]{Madotto2020AttentionOP}
Andrea Madotto, Zhaojiang Lin, Chien-Sheng Wu, Jamin Shin, and Pascale Fung.
  2020.
\newblock Attention over parameters for dialogue systems.
\newblock \emph{ArXiv}, abs/2001.01871.

\bibitem[{Majumder et~al.(2020)Majumder, Hong, Peng, Lu, Ghosal, Gelbukh,
  Mihalcea, and Poria}]{majumder-etal-2020-mime}
Navonil Majumder, Pengfei Hong, Shanshan Peng, Jiankun Lu, Deepanway Ghosal,
  Alexander Gelbukh, Rada Mihalcea, and Soujanya Poria. 2020.
\newblock \href {https://doi.org/10.18653/v1/2020.emnlp-main.721} {{MIME}:
  {MIM}icking emotions for empathetic response generation}.
\newblock In \emph{Proceedings of the 2020 Conference on Empirical Methods in
  Natural Language Processing (EMNLP)}, pages 8968--8979, Online. Association
  for Computational Linguistics.

\bibitem[{Moryossef et~al.(2019)Moryossef, Goldberg, and
  Dagan}]{moryossef-etal-2019-step}
Amit Moryossef, Yoav Goldberg, and Ido Dagan. 2019.
\newblock \href {https://doi.org/10.18653/v1/N19-1236} {{S}tep-by-step:
  {S}eparating planning from realization in neural data-to-text generation}.
\newblock In \emph{Proceedings of the 2019 Conference of the North {A}merican
  Chapter of the Association for Computational Linguistics: Human Language
  Technologies, Volume 1 (Long and Short Papers)}, pages 2267--2277,
  Minneapolis, Minnesota. Association for Computational Linguistics.

\bibitem[{Novikova et~al.(2017)Novikova, Du{\v{s}}ek, Cercas~Curry, and
  Rieser}]{novikova-etal-2017-need}
Jekaterina Novikova, Ond{\v{r}}ej Du{\v{s}}ek, Amanda Cercas~Curry, and Verena
  Rieser. 2017.
\newblock \href {https://doi.org/10.18653/v1/D17-1238} {Why we need new
  evaluation metrics for {NLG}}.
\newblock In \emph{Proceedings of the 2017 Conference on Empirical Methods in
  Natural Language Processing}, pages 2241--2252, Copenhagen, Denmark.
  Association for Computational Linguistics.

\bibitem[{Oraby et~al.(2019)Oraby, Harrison, Ebrahimi, and
  Walker}]{oraby_curate_2019}
Shereen Oraby, Vrindavan Harrison, Abteen Ebrahimi, and Marilyn Walker. 2019.
\newblock \href {https://doi.org/10.18653/v1/P19-1596} {Curate and {Generate}:
  {A} {Corpus} and {Method} for {Joint} {Control} of {Semantics} and {Style} in
  {Neural} {NLG}}.
\newblock In \emph{Proceedings of the 57th {Annual} {Meeting} of the
  {Association} for {Computational} {Linguistics}}, pages 5938--5951, Florence,
  Italy. Association for Computational Linguistics.

\bibitem[{Oraby et~al.(2018)Oraby, Reed, Tandon, T.S., Lukin, and
  Walker}]{oraby_controlling_2018}
Shereen Oraby, Lena Reed, Shubhangi Tandon, Sharath T.S., Stephanie Lukin, and
  Marilyn Walker. 2018.
\newblock \href {https://doi.org/10.18653/v1/W18-5019} {Controlling
  {Personality}-{Based} {Stylistic} {Variation} with {Neural} {Natural}
  {Language} {Generators}}.
\newblock In \emph{Proceedings of the 19th {Annual} {SIGdial} {Meeting} on
  {Discourse} and {Dialogue}}, pages 180--190, Melbourne, Australia.
  Association for Computational Linguistics.

\bibitem[{Papineni et~al.(2002)Papineni, Roukos, Ward, and
  Zhu}]{papineni-etal-2002-bleu}
Kishore Papineni, Salim Roukos, Todd Ward, and Wei-Jing Zhu. 2002.
\newblock \href {https://doi.org/10.3115/1073083.1073135} {{B}leu: a method for
  automatic evaluation of machine translation}.
\newblock In \emph{Proceedings of the 40th Annual Meeting of the Association
  for Computational Linguistics}, pages 311--318, Philadelphia, Pennsylvania,
  USA. Association for Computational Linguistics.

\bibitem[{Pavlick and Tetreault(2016)}]{pavlick_empirical_2016}
Ellie Pavlick and Joel Tetreault. 2016.
\newblock \href {https://doi.org/10.1162/tacl_a_00083} {An {Empirical}
  {Analysis} of {Formality} in {Online} {Communication}}.
\newblock \emph{Transactions of the Association for Computational Linguistics},
  4:61--74.

\bibitem[{Peng et~al.(2020)Peng, Zhu, Li, Li, Li, Zeng, and
  Gao}]{Peng2020FewshotNL}
Baolin Peng, Chenguang Zhu, Chunyuan Li, Xiujun Li, Jinchao Li, Michael Zeng,
  and Jianfeng Gao. 2020.
\newblock Few-shot natural language generation for task-oriented dialog.
\newblock In \emph{EMNLP}.

\bibitem[{Peng et~al.(2018)Peng, Ghazvininejad, May, and
  Knight}]{peng_towards_2018}
Nanyun Peng, Marjan Ghazvininejad, Jonathan May, and Kevin Knight. 2018.
\newblock \href {https://doi.org/10.18653/v1/W18-1505} {Towards {Controllable}
  {Story} {Generation}}.
\newblock In \emph{Proceedings of the {First} {Workshop} on {Storytelling}},
  pages 43--49, New Orleans, Louisiana. Association for Computational
  Linguistics.

\bibitem[{Radford et~al.(2019)Radford, Wu, Child, Luan, Amodei, and
  Sutskever}]{radford_language_nodate}
Alec Radford, Jeff Wu, R.~Child, David Luan, Dario Amodei, and Ilya Sutskever.
  2019.
\newblock Language models are unsupervised multitask learners.

\bibitem[{Rastogi et~al.(2019)Rastogi, Zang, Sunkara, Gupta, and
  Khaitan}]{rastogi2019towards}
Abhinav Rastogi, Xiaoxue Zang, Srinivas Sunkara, Raghav Gupta, and Pranav
  Khaitan. 2019.
\newblock Towards scalable multi-domain conversational agents: The
  schema-guided dialogue dataset.
\newblock \emph{arXiv preprint arXiv:1909.05855}.

\bibitem[{See et~al.(2019)See, Roller, Kiela, and Weston}]{see_what_2019}
Abigail See, Stephen Roller, Douwe Kiela, and Jason Weston. 2019.
\newblock \href {https://doi.org/10.18653/v1/N19-1170} {What makes a good
  conversation? {How} controllable attributes affect human judgments}.
\newblock In \emph{Proceedings of the 2019 {Conference} of the {North}
  {American} {Chapter} of the {Association} for {Computational} {Linguistics}:
  {Human} {Language} {Technologies}, {Volume} 1 ({Long} and {Short} {Papers})},
  pages 1702--1723, Minneapolis, Minnesota. Association for Computational
  Linguistics.

\bibitem[{Sheng et~al.(2019)Sheng, Chang, Natarajan, and
  Peng}]{sheng-etal-2019-woman}
Emily Sheng, Kai-Wei Chang, Premkumar Natarajan, and Nanyun Peng. 2019.
\newblock \href {https://doi.org/10.18653/v1/D19-1339} {The woman worked as a
  babysitter: On biases in language generation}.
\newblock In \emph{Proceedings of the 2019 Conference on Empirical Methods in
  Natural Language Processing and the 9th International Joint Conference on
  Natural Language Processing (EMNLP-IJCNLP)}, pages 3407--3412, Hong Kong,
  China. Association for Computational Linguistics.

\bibitem[{Socher et~al.(2013)Socher, Perelygin, Wu, Chuang, Manning, Ng, and
  Potts}]{socher-etal-2013-recursive}
Richard Socher, Alex Perelygin, Jean Wu, Jason Chuang, Christopher~D. Manning,
  Andrew Ng, and Christopher Potts. 2013.
\newblock \href {https://www.aclweb.org/anthology/D13-1170} {Recursive deep
  models for semantic compositionality over a sentiment treebank}.
\newblock In \emph{Proceedings of the 2013 Conference on Empirical Methods in
  Natural Language Processing}, pages 1631--1642, Seattle, Washington, USA.
  Association for Computational Linguistics.

\bibitem[{Sun et~al.(2021)Sun, Moon, Crook, Roller, Silvert, Liu, Wang, Liu,
  Cho, and Cardie}]{sun-etal-2021-adding}
Kai Sun, Seungwhan Moon, Paul Crook, Stephen Roller, Becka Silvert, Bing Liu,
  Zhiguang Wang, Honglei Liu, Eunjoon Cho, and Claire Cardie. 2021.
\newblock \href {https://doi.org/10.18653/v1/2021.naacl-main.124} {Adding
  chit-chat to enhance task-oriented dialogues}.
\newblock In \emph{Proceedings of the 2021 Conference of the North American
  Chapter of the Association for Computational Linguistics: Human Language
  Technologies}, pages 1570--1583, Online. Association for Computational
  Linguistics.

\bibitem[{Vaswani et~al.(2017)Vaswani, Shazeer, Parmar, Uszkoreit, Jones,
  Gomez, Kaiser, and Polosukhin}]{NIPS2017_3f5ee243}
Ashish Vaswani, Noam Shazeer, Niki Parmar, Jakob Uszkoreit, Llion Jones,
  Aidan~N Gomez, \L~ukasz Kaiser, and Illia Polosukhin. 2017.
\newblock \href
  {https://proceedings.neurips.cc/paper/2017/file/3f5ee243547dee91fbd053c1c4a845aa-Paper.pdf}
  {Attention is all you need}.
\newblock In \emph{Advances in Neural Information Processing Systems},
  volume~30. Curran Associates, Inc.

\bibitem[{Verma and Srinivasan(2019)}]{verma_lexical_2019}
Gaurav Verma and Balaji~Vasan Srinivasan. 2019.
\newblock \href {http://arxiv.org/abs/1909.08349} {A {Lexical}, {Syntactic},
  and {Semantic} {Perspective} for {Understanding} {Style} in {Text}}.
\newblock \emph{arXiv:1909.08349 [cs]}.
\newblock ArXiv: 1909.08349.

\bibitem[{Wang et~al.(2021)Wang, Hsu, Wu, and Yang}]{9362067}
Yi-Hsuan Wang, Jia-Hao Hsu, Chung-Hsien Wu, and Tsung-Hsien Yang. 2021.
\newblock \href {https://doi.org/10.1109/ISCSLP49672.2021.9362067}
  {Transformer-based empathetic response generation using dialogue situation
  and advanced-level definition of empathy}.
\newblock In \emph{2021 12th International Symposium on Chinese Spoken Language
  Processing (ISCSLP)}, pages 1--5.

\bibitem[{Wen et~al.(2015)Wen, Ga{\v{s}}i{\'c}, Mrk{\v{s}}i{\'c}, Su, Vandyke,
  and Young}]{wen-etal-2015-semantically}
Tsung-Hsien Wen, Milica Ga{\v{s}}i{\'c}, Nikola Mrk{\v{s}}i{\'c}, Pei-Hao Su,
  David Vandyke, and Steve Young. 2015.
\newblock \href {https://doi.org/10.18653/v1/D15-1199} {Semantically
  conditioned {LSTM}-based natural language generation for spoken dialogue
  systems}.
\newblock In \emph{Proceedings of the 2015 Conference on Empirical Methods in
  Natural Language Processing}, pages 1711--1721, Lisbon, Portugal. Association
  for Computational Linguistics.

\bibitem[{Yu et~al.(2017)Yu, Zhang, Wang, and Yu}]{Yu2017SeqGANSG}
Lantao Yu, W.~Zhang, J.~Wang, and Y.~Yu. 2017.
\newblock Seqgan: Sequence generative adversarial nets with policy gradient.
\newblock In \emph{AAAI}.

\bibitem[{Zhang et~al.(2020)Zhang, Sax, Zamir, Guibas, and
  Malik}]{Zhang2020SideTuningAB}
J.~Zhang, Alexander Sax, A.~Zamir, L.~Guibas, and J.~Malik. 2020.
\newblock Side-tuning: A baseline for network adaptation via additive side
  networks.
\newblock In \emph{ECCV}.

\bibitem[{Zhang et~al.(2018)Zhang, Guo, Fan, Lan, Xu, and
  Cheng}]{zhang_learning_2018}
Ruqing Zhang, Jiafeng Guo, Yixing Fan, Yanyan Lan, Jun Xu, and Xueqi Cheng.
  2018.
\newblock \href {https://doi.org/10.18653/v1/P18-1102} {Learning to {Control}
  the {Specificity} in {Neural} {Response} {Generation}}.
\newblock In \emph{Proceedings of the 56th {Annual} {Meeting} of the
  {Association} for {Computational} {Linguistics} ({Volume} 1: {Long}
  {Papers})}, pages 1108--1117, Melbourne, Australia. Association for
  Computational Linguistics.

\bibitem[{Zhou and Wang(2018)}]{zhou-wang-2018-mojitalk}
Xianda Zhou and William~Yang Wang. 2018.
\newblock {M}oji{T}alk: Generating emotional responses at scale.
\newblock In \emph{Proceedings of the 56th Annual Meeting of the Association
  for Computational Linguistics (Volume 1: Long Papers)}.

\bibitem[{Ziegler et~al.(2019)Ziegler, Stiennon, Wu, Brown, Radford, Amodei,
  Christiano, and Irving}]{Ziegler2019FineTuningLM}
Daniel~M. Ziegler, Nisan Stiennon, Jeffrey Wu, T.~Brown, A.~Radford, Dario
  Amodei, Paul Christiano, and Geoffrey Irving. 2019.
\newblock Fine-tuning language models from human preferences.
\newblock \emph{ArXiv}, abs/1909.08593.

\end{thebibliography}
	
	\clearpage
	\appendix
\section{Statistics of Style Annotation for SGD Data}
\label{append:data-stats}

\begin{table}[!h]
    \footnotesize
    \centering
    \resizebox{.48\textwidth}{!}{
    \begin{tabulary}{.48\textwidth}{ LRRR }
        \toprule
        & \textbf{Train} & \textbf{Dev} & \textbf{Test} \\
        \midrule
        System Turn & 164,982 & 24,363 & 42,297 \\
        Meaning Representations & 1,698 & 721 & 1,137 \\
        Templates & 118,715 & 19,963 & 34,598 \\
        \bottomrule
    \end{tabulary}}
    \caption{\label{table:sgd-data-stats} Schema-Guided Dialogue dataset statistics.}
\end{table}

\paragraph{Normalized inverse document frequency calculation}
For a word $w$ in a template $d$, the Inverse Document Frequency (IDF) of $w$ is
\begin{align}
    \text{IDF}(w) = \log \Bigg(\frac{N}{1 + \big|\{d \in D: w \in d\}\big|} \Bigg)
\end{align}
where $N$ is the total number of templates in the training data $D$, and $|\{d \in D: w \in d\}|$ is the number of those templates $d$ in the training data that contain $w$. The Normalized IDF (NIDF) is obtained via the min-max normalization
\begin{align}
    \text{NIDF}(w) = \frac{\text{IDF}(w) - \min\limits_{w' \in D}(\text{IDF}(w'))}{\max\limits_{w' \in D}(\text{IDF}(w')) - \min\limits_{w' \in D}(\text{IDF}(w'))}
\end{align}
where $\max\limits_{w' \in D}(\text{IDF}(w'))$ and $\min\limits_{w' \in D}(\text{IDF}(w'))$ are the maximum and minimum IDF value of all the words in the training data. We use the maximum NIDF of all the words in a template to determine whether the template contains rarely used words.

\paragraph{Semantic style parameter annotation details}
The formality corpus\footnote{\url{http://www.seas.upenn.edu/~nlp/resources/formality-corpus.tgz}} \citep{pavlick_empirical_2016}, which contains sentence-level formality annotations, is used for training the formality classifier. The Stanford sentiment treebank (SST)\footnote{\url{https://nlp.stanford.edu/sentiment/}} \citep{socher-etal-2013-recursive} is used for training the sentiment classifier. The SST data consists of 5 classes -- \textit{very negative}, \textit{negative}, \textit{neutral}, \textit{positive}, and \textit{very positive}. We combine ``\textit{very negative}'' and ``\textit{negative}'' into one class and ``\textit{very positive}'' and ``\textit{positive}'' into one class. Finally, the empathetic reactions data\footnote{\url{https://github.com/wwbp/empathic_reactions}} \citep{Buechel18emnlp} is used for training the empathy classifier. We use only the \textit{empathy} label from the data.

\begin{table}[!h]
    \footnotesize
    \centering
    \resizebox{.48\textwidth}{!}{
    \begin{tabularx}{.48\textwidth}{ lX }
        \toprule
        \textbf{Style parameters} & \textbf{Dataset} \\
        \midrule
         Formal & Formality corpus \\
         \midrule
         Positive & \multirow{2}{\linewidth}{Stanford sentiment treebank (SST)} \\
         Negative & \\
         \midrule
         Empathy & Empathetic reactions \\
        \bottomrule
    \end{tabularx}}
    \caption{\label{table:semantic-styles} Semantic style parameters with their possible values and associated training datasets.}
\end{table}

\begin{table}[!h]
    \footnotesize
    \centering
    \resizebox{.48\textwidth}{!}{
    \begin{tabulary}{\textwidth}{ LLCCC }
        \toprule
        \textbf{Category} & \textbf{Style} & \textbf{Train} & \textbf{Dev} & \textbf{Test} \\
         \midrule
         
         \multirow{7}{*}{Lexical styles} & Short & 31.42\% & 36.96\% & 32.23\% \\
         \cmidrule{2-5}
         
         & Long & 26.49\% & 25.13\% & 25.44\% \\
         \cmidrule{2-5}
         
         & Has rare word & 35.83\% & 37.76\% & 42.04\% \\
         \cmidrule{2-5}
         
         & First person pron. & 28.48\% & 28.88\% & 28.75\% \\
         \cmidrule{2-5}
         
         & Second person pron.& 52.89\% & 52.50\% & 60.18\% \\
         \cmidrule{2-5}

         & Descriptive & 5.00\% & 5.32\% & 3.88\% \\
         \midrule
         
         \multirow{5}{*}{Semantic styles}& Formal & 30.59\% & 34.27\% & 29.76\% \\
         \cmidrule{2-5}
         
         & Negative & 6.55\% & 4.33\% & 4.95\% \\
         \cmidrule{2-5}
         
         & Positive & 39.70\% & 42.43\% & 37.41\% \\
         \cmidrule{2-5}
         
         & Empathy & 27.72\% & 30.04\% & 25.65\% \\
        \bottomrule
    \end{tabulary}}
    \caption{\label{table:data-distribution} Data distribution for all of the style parameters we annotate in SGD.}
\end{table}
\begin{table*}[!h]
    \centering
    \footnotesize
    \begin{tabularx}{\textwidth}{ Xccc }
        \toprule
        \textbf{MR:} \verb|OFFER(restaurant_name=$slot1), OFFER(city=$slot2)| & \textbf{$\gamma_{gm}$} & \textbf{$\alpha$} & \textbf{$\lambda$} \\
        \midrule
        
        \verb|$slot1| is a nice restaurant in \verb|$slot2|. & 0.1 & 0.01 & 1 \\
        \verb|$slot1| is a nice restaurant in \verb|$slot2|. & 0.1 & 0.1 & 1 \\
        \verb|$slot1| is a good restaurant in \verb|$slot2|. & 0.1 & 1 & 1 \\
        \cmidrule{1-4}
        
        \verb|$slot1| is a nice restaurant in \verb|$slot2| & 0.3 & 0.01 & 0.01 \\
        \verb|$slot1| is a restaurant in \verb|$slot2|. & 0.3 & 0.01 & 0.1 \\
        \verb|$slot1| is a restaurant in \verb|$slot2|. & 0.3 & 0.01 & 1 \\
        \cmidrule{1-4}
        
        There's the \verb|$slot4| \verb|$slot1| restaurant in \verb|$slot2|. Do you like the same restaurant. Do you like this one more? & 0.5 & 0.01 & 0.01 \\
        \verb|$slot1| \verb|$slot2| I have a restaurant has good rating restaurant has \verb|$slot1| & 0.5 & 0.01 & 0.1 \\
        \verb|$slot1| \verb|$slot2| I have a restaurant has good rating restaurant has a restaurant name is & 0.5 & 0.01 & 1 \\
        \cmidrule{1-4}

        \verb|$slot1| is a nice restaurant in \verb|$slot2|. Do i suggest you like that one I've found? You can have a nice meal. & 0.5 & 0.1 & 1 \\
        I recommend mathemat, a	\verb|$slot1| is in \verb|$slot2|. & 0.5 & 1 & 1 \\
        \cmidrule{1-4}
        
        \verb|$slot4| \verb|$slot4| \verb|$slot4| there is \verb|$slot4| there is & 0.9 & 0.01 & 1\\ 
        \bottomrule
    \end{tabularx}
    \caption{\label{table:pplm-hyperparameters} Example outputs of PPLM for controlling ``\textit{positive sentiment}'' with various hyper-parameters. Increasing $\gamma_{gm}$ and $\alpha$ encourages style control but can lead to ungrammatical outputs or run into degeneration. Increasing $\lambda$ encourages sentence fluency and semantic accuracy; however, careful hyper-parameters fine-tuning is required to ensure reasonable output quality.}
\end{table*}

\section{Hyper-parameters Tuning for PPLM}
\label{append:hyper-parameters}

There are several hyper-parameters in PPLM models. We refer the readers to \citet{Dathathri2020PlugAP} for the full model details. From Section \ref{subsec:guided-fine-tune}, the PPLM model can be written as an optimization problem of the form:
\begin{align}
    \min_{\Delta H_t} \quad & L_{\text{CE}}\Big(f \Big( \sum_{t=1}^T \widetilde{o}_t / T \Big) \Big) + \lambda D_{\text{KL}}(\widetilde{o}_t\|o_t) \\
    \text{s.t.} \quad & \widetilde{o}_t = \text{LM}(x_t, \widetilde{H}_t) \\
    & \widetilde{H}_t = H_t + \Delta H_t \\
    & o_t = \text{LM}(x_t, H_t)
\end{align}
where $\lambda$ is a hyper-parameter that scales the KL divergence and $f(\cdot)$ is the semantic style classifier learned to produce $p(a|x)$. We minimize the cross-entropy loss $L_{\text{CE}}(\cdot)$ of the attribute model and the Kullback-Leibler (KL) divergence $D_{\text{KL}}(\cdot \| \cdot)$ of the language model. For each time step $t$, the GPT-2 model generates a history of weight matrices $H_t$ for all hidden layers. To control the generation, we shift the weight matrices $H_t$ in the direction that accomplishes two goals: 1) minimize cross-entropy (CE) of the attribute $a$ under the conditional attribute model $p(a|x)$ and 2) minimize the KL divergence between itself and the unmodified language model $p(x)$. Let $\Delta H_t$ be the small shift of $H_t$ such that $\widetilde{H}_t = H_t + \Delta H_t$ can shift the last layer (logit vector) to achieve the above two goals. The logit vector $o_t$ is obtained by a forward pass of the LM, \emph{i.e.} $o_t = \text{LM}(x_t, H_t)$. The next generated token $x_{t+1}$ is sampled as $x_{t+1} \sim \text{softmax}(W o_t)$ where $W$ is a linear transformation that maps $o_t$ to a probability distribution of vocabulary size. 

The shift $\Delta H_t$ is then updated with gradient descent as follows:
\begin{align}\label{eq:gradient-descent}
    \Delta H_t \leftarrow &\Delta H_t - \\
    &\alpha \Big( \nabla_{\Delta H_t} L_{\text{CE}}(\cdot) + \lambda \nabla_{\Delta H_t} D_{\text{KL}}(\cdot\|\cdot) \Big) \nonumber
\end{align}
where $\alpha$ is the step size and the update can be repeated until it converges or up to a certain time step to obtain the shift $\Delta H_t$. At generation time, a \textit{post-norm fusion} that samples the next token $x_{t+1}$ based on the shifted distribution $\widetilde{p}_{t+1}$ and the unmodified distribution $p_{t+1}$ is done to increase the language fluency:
\begin{align}
    x_{t+1} \sim \frac{1}{\beta} \Big( \widetilde{p}_{t+1}^{\gamma_{gm}} \, p_{t+1}^{1-\gamma_{gm}} \Big)
\end{align}
$\beta$ is a normalizing factor such that it forms a valid distribution. As $\gamma_{gm} \rightarrow 1$ this converges to the distribution from the shifted LM, and as $\gamma_{gm} \rightarrow 0$ it converges to the unmodified unconditional LM distribution. We observe that the hyper-parameters $\lambda, \alpha, \gamma_{gm}$ can be tuned to affect the generation. In practice, we found that increasing $\gamma_{gm}$ led to nonsensical results very quickly as the generated text is no longer conformed to the given meaning representation. A larger step size $\alpha$ also led to unstable and nonsensical results compared to a smaller step size as it moves $H_t$ further away from its original position. In practice, a larger $\lambda$ helps keep the fluency and the semantic of the generated text. Nevertheless, its effect is less influential than $\alpha$ and $\gamma_{gm}$. Table \ref{table:pplm-hyperparameters} illustrates the effect of hyper-parameters on the model outputs.
\begin{table*}[!h]
    \centering
    \resizebox{.8\textwidth}{!}{
    \begin{tabulary}{\textwidth}{ ll }
        \toprule
        \multicolumn{2}{l}{\textbf{MR:} \texttt{OFFER(address=\$slot1), OFFER(rating=\$slot2)}} \\
        \midrule
        W/o Style Control & There is a nice house at \verb|$slot1| with a \verb|$slot2| rating. \\
        \midrule
        Short & \verb|$slot1| is rated \verb|$slot2|. \\
        \midrule        
        Long & There is a house at \verb|$slot1| with a rating of \verb|$slot2|. Would you like that one? \\
        \midrule
        
        %W/o Style Control &  There is a nice house at \verb|$slot1| with a \verb|$slot2| rating. \\
        Has Rare Word & There is a lovely residence located at \verb|$slot1|. It has a rating of \verb|$slot2|. \\
        \midrule
        
        %W/o Style Control &  There is a nice house at \verb|$slot1| with a \verb|$slot2| rating. \\
        First Person Pron. & I found a house at \verb|$slot1| with a \verb|$slot2| rating. \\
        \midrule
        
        %W/o Style Control &  There is a nice house at \verb|$slot1| with a \verb|$slot2| rating. \\
        Second Person Pron. & There is a house at \verb|$slot1| with a \verb|$slot2| rating. Would you like that one? \\
        \midrule
        
       % W/o Style Control &  There is a nice house at \verb|$slot1| with a \verb|$slot2| rating. \\
        Descriptive & There is a nice house with a \verb|$slot2| rating available at \verb|$slot1|.\\
        \bottomrule \\
        
        \toprule
        \multicolumn{2}{l}{\textbf{MR:} \texttt{OFFER\_INTENT(intent=intent)}} \\
        \midrule
        W/o Style Control & Would you like to make a reservation for this house? \\
        \midrule
        Short & Do you want to reserve it? \\
        \midrule        
        Long & Do you want me to go ahead and reserve a room at the hotel? \\
        \midrule
        
        %W/o Style Control &  Would you like to reserve a room at the hotel? \\
        Has Rare Word & Would you like to continue with the hotel reservation? \\
        \midrule
        
        %W/o Style Control &  Would you like to make a reservation for this house? \\
        First Person Pron. & Shall I book the rooms in that hotel now? \\
        \midrule
        
        %W/o Style Control & Shall I book the rooms in that hotel now? \\
        Second Person Pron. & Would you like to make a reservation for this house? \\
        \midrule
        
        %W/o Style Control & Would you like to make a reservation for this house? \\
        Descriptive & Would you like to reserve a room at this hotel? \\
        \bottomrule \\
        
    \end{tabulary}}
    \caption{\label{table:additional-ct-outputs} Example outputs using conditional training for the six lexical styles described in Section \ref{sec:data}.}% The examples show that conditional training is effective in controlling the lexical styles except for ``\textit{descriptive}''.}
\end{table*}

\section{Additional Experimental Results}
\label{append:additional-experiments}
We include additional example outputs for lexical styles in Table \ref{table:additional-ct-outputs}.

\paragraph{Multiple Styles Control}
\label{append:multiple-styles}
In this section, we demonstrate the possibility of controlling multiple styles with the methods described in Section \ref{sec:methods}. To control multiple styles with conditional training, we condition the CT model on multiple control variables $P(x|a_1, \cdots, a_n)$. The control variables are concatenated and then added to the input as special tokens after the \verb|[BOS]| token. To control multiple styles with PPLM, we simply add the cross-entropy loss of each semantic style classifier to the objective function. Extra hyper-parameters can be introduced to control the significance of each style. Similarly, for (beam search) weighted decoding, the distribution is re-weighted by
\begin{align}
    p(w|a_1, \cdots, a_n) = \text{Softmax} \Big(p(w) \prod_{i=1}^n \lambda_i p(a_i|w) \Big)
\end{align}
where $\lambda_i$ are hyper-parameters that allow us to determine the significance of each style $a_i$. Finally, conditional training and (beam search) weighted decoding can be used simultaneously by training a CT model and then applying WD or BSWD on the trained CT model during decoding time. Example outputs and the style accuracy of multiple style control using CT and BSWD are shown in Tables \ref{table:multiple-style-results} and \ref{table:ct-bswd-example}.

\begin{table}[h]
    \centering
    \footnotesize
    \begin{tabulary}{.48\textwidth}{LLLL}
        \toprule
        \textbf{Lexical Style} & \textbf{Lexical Sty. Acc.} & \textbf{Semantic Style} & \textbf{Semantic Sty. Acc.} \\
        \midrule
        Short & 55.4\% & Formal & 2.0\% \\
        Long & 95.2\% & Formal & 39.1\% \\
        \midrule
        Short & 69.2\% & Positive & 20.3\% \\
        Long & 84.4\% & Positive & 44.8\% \\
        \bottomrule
    \end{tabulary}
    \caption{\label{table:multiple-style-results} Style accuracy of multiple-style control using CT and BSWD with beam width $B=2$. The parameter \textit{length} is controlled by CT and the parameters \textit{formality} and \textit{sentiment} are controlled by the BSWD.}
\end{table}

\begin{table*}[!h]
    \centering
    \footnotesize
    \begin{tabularx}{\textwidth}{ Xll }
        \toprule
        \textbf{MR:} \verb|OFFER(restaurant_name=$slot), OFFER(city=$slot2)| & \textbf{Lexical Sty.} & \textbf{Semantic Sty.} \\
        \midrule
        
        \verb|$slot1| in \verb|$slot2| is a very popular indian restaurant. & \multirow{2}{*}{W/o Style Control} & \multirow{2}{*}{W/o Style Control} \\
        \verb|$slot1| is a restaurant in \verb|$slot2| that serves pizza and pasta. \\
        \cmidrule{1-3}
        
        \verb|$slot1| is a restaurant in \verb|$slot2|. & \multirow{2}{*}{Short} & \multirow{2}{*}{W/o Style Control} \\
        \verb|$slot1| serves italian food in \verb|$slot2|. \\
        \cmidrule{1-3}
        
        Okay! The \verb|$slot1| located in \verb|$slot2|, as for the tofu, it has a pretty good rating. Would you like to try for a lunch or dinner reservation? & \multirow{4}{*}{Long} & \multirow{4}{*}{W/o Style Control} \\
        Okay! The \verb|$slot1| located in \verb|$slot2|, as for the tofu, it has a pretty good rating. Would you like to try for a meal there? \\
        \midrule
        
        \verb|$slot1| is a nice restaurant located in \verb|$slot2|. & \multirow{2}{*}{W/o Style Control} & \multirow{2}{*}{Positive} \\
        \verb|$slot1| might interest you, and it is located in \verb|$slot2|. \\
        \cmidrule{1-3}
        
        \verb|$slot1| is a good restaurant in \verb|$slot2|. & \multirow{2}{*}{Short} & \multirow{2}{*}{Positive} \\
        \verb|$slot1| restaurant is also in \verb|$slot2|. \\ 
        \cmidrule{1-3}
        
        Okay! The restaurant \verb|$slot1| located right inside \verb|$slot2| is a good one and it has many vegetarian side dishes. & \multirow{4}{*}{Long} & \multirow{4}{*}{Positive} \\
        Okay! The restaurant \verb|$slot1| located right inside \verb|$slot2| is a good one. Do you wish to have your lunch here? \\
        \bottomrule \\

        \toprule
        \textbf{MR:} \verb|OFFER(address=$slot1), OFFER(rating=$slot2)| & \textbf{Lexical Sty.} & \textbf{Semantic Sty.} \\
        \midrule
        
        There is a \verb|$slot2| rated house at \verb|$slot1|. & \multirow{2}{*}{W/o Style Control} & \multirow{2}{*}{W/o Style Control} \\
        There is a house at \verb|$slot1| with a \verb|$slot2| rating. \\
        \cmidrule{1-3}
        
        \verb|$slot1| has a \verb|$slot2| rating & \multirow{2}{*}{Short} & \multirow{2}{*}{W/o Style Control} \\
        \verb|$slot1| with a \verb|$slot2| rating \\ 
        \cmidrule{1-3}
        
        There is a house at \verb|$slot1| with a \verb|$slot2| rating. would you like to stay here?  & \multirow{3}{*}{Long} & \multirow{3}{*}{W/o Style Control} \\
        There is a house at \verb|$slot1| that you might be interested in. It has a \verb|$slot2| average rating. \\
        \midrule
        
        There is a lovely house at \verb|$slot1| with a rating of \verb|$slot2| & \multirow{2}{*}{W/o Style Control} & \multirow{2}{*}{Positive} \\
        There is a good house at \verb|$slot1| with a rating of \verb|$slot2| \\
        \cmidrule{1-3}
        
        \verb|$slot1| has a rating of \verb|$slot2| & \multirow{2}{*}{Short} & \multirow{2}{*}{Positive} \\
        \verb|$slot1| with rating of \verb|$slot2| \\
        \cmidrule{1-3}
        
        There is a lovely residence located at \verb|$slot1| with a rating of \verb|$slot2| & \multirow{2}{*}{Long} & \multirow{2}{*}{Positive} \\
        There is a nice house located at \verb|$slot1| with a rating of \verb|$slot2| \\
        
        \bottomrule
    \end{tabularx}
    \caption{\label{table:ct-bswd-example} Example outputs by combining CT and BSWD with beam width $B=2$. Lexical style is controlled by CT and semantic style is controlled by the BSWD. Note how the lexical style "long" tends to yield outputs that include more hallucinated content in an attempt to fulfill the required style goal(s).}
\end{table*}

\newpage
\section{Human Evaluation Design}
\label{append:human-eval}

The goal of the annotation task is to evaluate the style of the automatically generated templates that express specific information to the user. The templates follow a certain style value such as ``\textit{formal}''. The annotators are asked to imagine that they are having a conversation with the system, and that the template presented to them is an example of something that the system may say to them. The template may be in the form of a question, statement, confirmation, etc. Table \ref{table:human-eval-example} illustrates two sample MRs and four generated templates, which we lexicalize with a map of values (one per slot) to make the annotation task more intuitive.

\begin{table*}[!h]
    \centering
    \footnotesize
    \begin{tabulary}{\textwidth}{ LlL }
        \toprule
        \textbf{Domain} & \textbf{Simplified MR} & \textbf{Output template} \\
        
        \midrule
        \multirow{2}{*}{Movies} & \multirow{2}{*}{\textsl{request city none}} & What is the location and type of movie you are interested in? \\
        \cmidrule{3-3}
        & \textsl{request movie\_name none} & What movie do you want to see and in what location? \\
        
        \midrule
        \multirow{3}{*}{Restaurants} & \multirow{2}{*}{\textsl{confirm restaurant\_name [Ala Romana]}} &  Please confirm: booking a table at [Ala Romana] on [March 1st]. \\
        \cmidrule{3-3}
        & \textsl{confirm date [March 1st]} &Please confirm the following: you want a table at [Ala Romana] on [March 1st]. \\
        
        \bottomrule
    \end{tabulary}
    \caption{\label{table:human-eval-example} Example human evaluation worksheet.}
\end{table*}

\end{document}